\def\year{2022}\relax
\documentclass[letterpaper]{article} 
\usepackage{aaai22}  
\usepackage{times}  
\usepackage{helvet}  
\usepackage{courier}  
\usepackage[hyphens]{url}  
\usepackage{graphicx} 
\usepackage{natbib}  
\usepackage{caption} 
\DeclareCaptionStyle{ruled}{labelfont=normalfont,labelsep=colon,strut=off} 
\usepackage{algorithm}
\usepackage{algorithmic}

\usepackage{newfloat}
\usepackage{listings}
\floatstyle{ruled}
\newfloat{listing}{tb}{lst}{}
\floatname{listing}{Listing}
\newcommand{\nth}[1]{#1^{\text{th}}}

\usepackage{dsfont}
\usepackage{amsmath}
\usepackage{graphics}
\usepackage{amssymb}

\title{Detecting model drift using polynomial relations}
\author{
Eliran Roffe, Samuel Ackerman, Orna Raz and Eitan Farchi
}
\affiliations{
    \textsuperscript{\rm 1}IBM Research, Haifa, Israel\\

    eliranr@il.ibm.com, samuel.ackerman@ibm.com
}

\begin{document}

\maketitle

\begin{abstract}
Machine learning models serve critical functions, such as classifying loan applicants as good or bad risks. Each model is trained under the assumption that the data used in training and in the field come from the same underlying unknown distribution. Often, this assumption is broken in practice. It is desirable to identify when this occurs, to minimize the impact on model performance.

We suggest a new approach to detecting change in the data distribution by identifying polynomial relations between the data features. We measure the strength of each identified relation using its R-square value.  A strong polynomial relation captures a significant trait of the data which should remain stable if the data distribution does not change. We thus use a set of learned strong polynomial relations to identify drift.  For a set of polynomial relations that are stronger than a given threshold, we calculate the amount of drift observed for that relation. The amount of drift is measured by calculating the Bayes Factor for the polynomial relation likelihood of the baseline data versus field data.  We empirically validate the approach by simulating a range of changes, and identify drift using the Bayes Factor of the polynomial relation likelihood change.
\end{abstract}



\section{Introduction}
Machine learning (ML) models may serve critical business functions, such as classifying loan applicants as good or bad risks, or classifying mammogram images as malignant or benign.
It is generally assumed that the data used in training, and the data used in field come from the same underlying unknown distribution. Often, this assumption is broken in practice, and in these cases the distribution change can cause the model's performance to degrade or change in some way.  It is therefore desirable to be able to identify when the data distribution has changed. In addition, it may be difficult to characterize the training data distribution in a way that is general but will still capture important aspects of the data, like complex relations between features.

The contribution of this paper is a new approach to detect change in the data distribution by identifying polynomial relations between the tabular structured data set features.  A relation consists of an estimated equation relating, say, features $X_1$ to $X_2$ and $X_3^2$, along with an estimation of the error, where $X_1$, $X_2$, and $X_3$ are features in the data set; these relations can be constructed by linear regression, for example.  Relations that are particularly strong (e.g., low error) on the baseline data can be said to characterize an aspect of that data. We then apply ML and statistical techniques (e.g., linear likelihood, Bayes Factor) over these relations to identify when they change significantly between the baseline and field data.  In this case, if a relation holds on the baseline data, and no longer applies when the model is deployed, it indicates a change in the data distribution (and may indicate a potential decrease  in the model performance).  Our contribution is an end-to-end way to identify the most relevant relations and then use these to identify distribution change.  

Several works capture challenges and best practices in detecting drift in the data distribution. In this work, we do not experiment with any example ML prediction models to verify that the model's performance changes due to distribution drift, since this will depend on the specific model used. Rather, we simply assume that the existence of significant distribution drift between a baseline and field data set has the potential to degrade the model performance; we consider this assumption highly intuitive without need for explicit demonstration.  Furthermore, regardless of their effect on model performance changes, changes in the data distribution, which we demonstrate that we can systematically detect, are important in their own right.

As a motivating example of distribution drift, consider a tabular data set where each observation is a cell phone model, and features include the phone's price and technical aspects and physical measurements of the phone. In a data set of phones from the pre-smartphone era, say there is a particularly strong correlation between the phone screen size and battery life.  Because smartphones and non-smartphones are fundamentally different in nature, the same relation may not hold between these features if the same data were collected today on a sample of mostly smartphones.  We would like to identify this kind of distribution change in data set features without domain knowledge; an ML model trained to, say, predict phone price based on other features, which was trained on the pre-smartphone data set, would likely have poor predictive performance today.  As another example, in the past, PHP and Javascript experience were important in predicting a programmer's salary.  Now, PHP is no longer crucial, so its effect on salary has decreased (or ceased to exist).

Our paper proceeds as follows: In Section~\ref{sec:related_work}, we mention some related work in drift detection which motivates this one.  In Section~\ref{sec:methodology}, we outline the methodology to construct polynomial feature relations and select the relevant subset of them, and to detect drift on a relation's strength using Bayes Factors.  In Section~\ref{sec:experiments}, we induce simulated drift of various degrees, and show that greater drift can be detected by higher values in the Bayes Factor in the strong relations we select.  These simulations (Section~\ref{ssec:experimentalDesign}) represent drift in the form of row and permutations to disrupt inter-feature correlations, or increasing `unfairness' in the form of changes in relations between features.  The results suggest that our method should be able to detect arbitrary forms of distribution drift, and that larger degrees of distortion should be easier to detect through their effect on the relations' Bayes Factors.  In Section~\ref{sec:discussion}, we discuss some challenges we intend to address in future work.  Section~\ref{sec:conclusion} concludes.

\section{Related Work} \label{sec:related_work}
Several works capture challenges and best practices in detecting drift in the data distribution. Some of these practical issues have been mentioned in earlier works (e.g., \cite{AFRZZ19}, \cite{ADFRZ20} and \cite{ARZ20}). 

These works tried to detect drift indirectly by non-parametrically testing the distribution of model prediction confidence for changes; that approach generalizes the method of drift detection and sidesteps domain-specific feature representation. For example, \cite{ADFRZ20} handled a subset of MNIST digits and then inserted drift (e.g., an unseen digit class) to deployment data in various settings (gradual/sudden changes in the drift proportion). A novel loss function was introduced to compare the performance (detection delay, Type-1 and 2 error rates) of a drift detector under different levels of drift class contamination.


In addition to these works on drift detection, there is extensive literature on identifying sets of data set features (i.e., columns or variables) that are strongly related in some way.  For instance, \cite{MAHall} and \cite{CBK} suggest correlation-based algorithms for feature selection, and \cite{HEPK} detects correlated columns in a relational database. These works and others rely on approaches to calculate correlation between features (e.g., Pearson or Spearman correlation, which measure the strength of linear and rank-order correlations, respectively), and approaches to measure the similarity between features (e.g., cosine similarity).  Furthermore, there is extensive literature on iterative selection of the best subset of data set features to form a linear model.

The work we present here is motivated by these works. We extend these earlier works by presenting a new automatic approach to identify potential relations between features in the data set which are not necessarily linear in these features (i.e., allow interactions and higher exponents). Then, we determine the influence of changes in a new data set on the strength of the `fit' of this relation.  Our work automates the whole process of identifying the changes in the data distribution, and does so in a more readable way of relations (which are human-interpretable), and a specific value which symbolizes the amount of calculated drift.

\section{Methodology}
\label{sec:methodology}

For the remainder of this paper, we assume we have a tabular structured data set of $n$ rows (observations) and $m$ columns (features), denoted $X_j,\:j=1,\dots,m$.  The features considered are all numeric (e.g., integer or real-valued) and not categorical.  We will typically assume the observations are mutually independent, and not time-indexed or spatially-related, such as a matrix representation of image pixel values.
We suggest a new approach for drift detection consisting of two main steps. The first step involves identifying relations between features and the strength of each relation. The second step consists of detecting drift by applying ML and statistical techniques (e.g., linear regression, linear likelihood, Bayes Factor) over these relations.

\subsection{Identifying relations between data set features\label{ssec:identifying_relations}}

We define a search space and a high-level outline of our proposed way to search it for statistically-significant relations. We proceed to implement this approach for the case of polynomial relations. Specifically, for a given feature $Y\in\{X_1,\dots,X_m\}$ in the data set, we consider all polynomial relations of $k$ other features and degree $\ell\in\{0,1,2,\dots\}$.  Without loss of generality, let the set $\mathbf{v}=\{X_1,\dots,X_k\}$ be a subset of $k$ of the $m$ features other than $Y$.  For a given $k$ and maximal degree $\ell$, define the polynomial embedding $\mathcal{P}$ of $k$ chosen features $\mathbf{v}$ as 

\[
\mathcal{P}\colon\:\{X_1,\dots,X_k\} \Rightarrow \left\{\prod_{j=1}^k X_{j}^{i_j}\right\}_{i_j \geq 0;\: \ell \geq \sum_{j=1}^{k}i_j}
\]

The embedding $\mathcal{P}$ generates a set of all possible candidate terms for inclusion in the polynomial that will keep it under degree $\ell$.  That is, all possible products (constant term, single features raised to exponents $1,\dots,\ell$, and multi-feature interactions) of the $k$ features raised to positive integer exponents, where the exponents in each product sum to at most $\ell$.  Exponents $i_j$ where $i_j=0$, which do not contribute to the sum, mean that feature $X_j$ is excluded.

For instance, for $k=\ell=2$, the values we choose for the remainder of the paper, $\mathbf{v}=\{X_1,\:X_2\}$, where $X_1$ and $X_2$ are any two different features. The polynomial embedding would be
$\mathcal{P}(\mathbf{v})=\mathcal{P}(\{X_1,\:X_2\})=\{1,\:X_1,\:X_2,\:X_1X_2,\:X_1^2,\:X_2^2\}$. For convenience, let $P_i,\:i=1,\dots,|\mathcal{P}(\mathbf{v})|$, denote the $\nth{i}$ term in the embedding. Here, for instance, $P_2=X_1$, and the exponent on $X_2$ is  $i_2=0$, so $X_2$ is ignored in this term.  

Furthermore, we can define a mapping template $\mathcal{M}\colon\:\mathds{R}^{|\mathcal{P}(\mathbf{v})|} \Rightarrow \mathds{R}$, which specifies a univariate real-valued output from real-valued vector embedding $\mathcal{P}$. In particular, the output of $\mathcal{M}$ will be a prediction of the value of $Y$ given the set of numeric predictors $\mathcal{P}(\mathbf{v})$, such as from an ML model.

We mention $\mathcal{M}$ to demonstrate that there may be a wide variety of methods to construct relations between $Y$ and other features.  Here, however, we restrict ourselves only to linear combinations, a particular form of the mapping template $\mathcal{M}$; these are denoted
$\mathcal{L}\colon\:\mathds{R}^{|\mathcal{P}(\mathbf{v})|} \Rightarrow \mathds{R}$, of the form

\[
\mathcal{L}\left(\mathcal{P}(\mathbf{v})\right) = \sum_{i=1}^{|\mathcal{P}(\mathbf{v})|} \beta_i P_i
\]

where $\beta_i\in\mathds{R}$
are term coefficients.  Since the elements $\{P_i\}$ themselves are maximal $\ell$-degree combinations of $k$ original data set features $\mathbf{v}=\{X_1,\dots,X_k\}$, any function of the form $\mathcal{L}$ above is a linear polynomial relation we will consider in this work.  
For instance, $\mathcal{L}$ could represent ordinary least-squares (OLS) linear regression (the technique we use in this paper), or Least Absolute Shrinkage and Selection Operator (LASSO, \cite{T1996}) or other forms of linear regression.  Since $\mathcal{L}$ is used to model the relationship between the feature $Y$ and the embedding features $\mathcal{P}(\mathbf{v})$, its output represents a prediction of $Y$'s value, so we can say that a linear polynomial relation takes the form $\hat{Y}=\mathcal{L}\left(\mathcal{P}(\mathbf{v})\right)$; the notation $\hat{Y}$ refers to an estimate or prediction of $Y$. 

Any particular choice of the form of $\mathcal{L}$ will give a particular set of values for the constants $\{\beta_i\}$ (e.g., LASSO will tend to set $\beta_i=0$ for non-significant regressors $P_i$) as well as a measurement of the error of the prediction $\hat{Y}$ from the true value $Y$.   For OLS linear regression, as noted in Section~\ref{ssec:BF_for_drift_detection}, this is done by estimating the variance of a Gaussian distributed error term, and by the R-squared coefficient of determination.  The $R^2$ value is $0\leq R^2\leq 1$, with higher values indicating that $\mathcal{L}\left(\mathcal{P}(\mathbf{v})\right)$ provides a more accurate prediction of $Y$, that is, that this relation is strong.

The only relations we consider here are polynomials of maximal degree $\ell=2$ using of up to $k=2$ other $m-1$ features for each target feature $Y$ out of the $m$ features; from these we retain only the strongest ones.  We note that it is likely that relations that are non-polynomial (e.g., using trigonometric functions or features raised to non-integer exponents), of higher degree than $\ell$ or involving more than $k=2$ other features, which may have stronger fit than the polynomial relations we use.  However, we restrict our search to these shorter polynomials for several reasons:
\begin{itemize}
    \item This limits the search space for relations; further, polynomials should be able to capture many observed relations and it is likely that allowing higher degrees or non-linear functions would lead to over-fitting or finding relations that may be artefacts. Limiting $k$ and $\ell$ further makes these relations more human-interpretable; that is, a human can verify that they are reasonable using basic domain knowledge.
    \item Though our algorithm finds strong relations, these are not the end goal in the themselves.  Ultimately, we use the selected relations as `sensors' which we will use to detect drift by observing changes in their fit, or `strength.'  Thus, to fulfill this goal, it is likely sufficient to find enough such strong relations, without worrying whether there may exist other, stronger relations without the constraints of linearity and $k$ and $\ell$.
    \item The polynomial framework, through linear regression, has a compact likelihood function, enabling the use of Bayes Factors (Section~\ref{ssec:BF_for_drift_detection}) to measure changes in fit.
\end{itemize}

We propose a basic algorithm to identify strong (good predictive fit) relations for a given feature $Y$, using up to $k$ of the $m-1$ other features in the data set.  Without restrictions on the size of the feature subset $k$ and the maximal degree $\ell$, the embedding search space $\mathcal{P}$ is theoretically of infinite size.  Furthermore, even with limits on $k$ and $\ell$, the infinite possibilities of constants $\{\beta_i\}$ make the space of linear polynomial relations $\{\mathcal{L}(\mathcal{P}(\mathbf{v}))\}$ infinite; however, choosing an algorithm such as linear regression leads us to choices of $\{\beta_i\}$ that actually provide good predictions of $Y$. As mentioned above, we use low values of both $k$ and $\ell$ to reduce over-fitting of the relations to each $Y$ and to make our heuristic search algorithm more efficient.

Our algorithm proceeds as follows:  Given a choice $k$, we form the $m\times m$ matrix of absolute values of pairwise correlations (e.g., Pearson) between the $m$ data set features. The following procedure is then repeated for each feature $X_j,\:j=1,\dots,m$, which each time is designated as the target $Y$: choose only the $k$ most-correlated (e.g., Pearson correlation) other features to $X_j$, and set these as $\mathbf{v}$, and then the embedding $\mathcal{P}(\mathbf{v})$ is determined.  We then form the linear relation $\hat{Y}=\mathcal{L}(\mathcal{P}(\mathbf{v}))$ and measure its $R^2$ value versus the true values $Y$. 

Only sufficiently `strong' relations, for which the $R^2\geq 0.9$, are retained as candidates for detecting drift, as we show.  The $R^2$ threshold of 0.9 was identified as an acceptable value in previous experiments; in addition to $\ell$ and $k$, it thus serves as a hyper-parameter which influences the number of relations identified, and can be set as an input to the algorithm.  

\subsection{Detecting drift using identified relations}
\label{ssec:BF_for_drift_detection}

Here we explore the utilization of a set of generated relations for distribution drift detection.  The likelihood function of a model or distribution is a function which takes as inputs the model parameters $\boldsymbol{\theta}$ and data inputs a matrix $\mathbf{X}$ (and possibly $\mathbf{Y}$ if it is a prediction model).  In that case, let $\mathbf{D}=(\mathbf{X},\mathbf{Y})$ denote a data set with the given pair.  Here, we use the general notation $\mathbf{X}$ and $\mathbf{Y}$ for inputs and output of a regression problem, but in our case, $\mathbf{X}=\begin{bmatrix}\mathcal{P}(\mathbf{v})\end{bmatrix}$ (the polynomial embedding of some other original data set of features) and $\mathbf{Y}$ is the target feature we are trying to relate linearly to $\mathcal{P}(\mathbf{v})$.  Thus, $m$, the number of columns in $\mathbf{X}$, is redefined here as $|\mathcal{P}(\mathbf{v})|$, the number of features in the embedding, and not the number of features in the original dataset to which $\mathcal{P}$ is applied. 

Assume $\mathbf{X}$ contains $n$ observations and $m$ feature columns.  Let $\mathbf{X}_j,\:j=1,\dots,m$ be the $\nth{j}$ column (feature) of the data matrix $\mathbf{X}$; if an intercept term is fit, $\mathbf{X}_1=\mathbf{1}$ (recall $P_1=1$), so $\beta_1$ (rather than $\beta_0$ as is typically used) corresponds to the intercept slope.
Let $X_{i,j},\: i=1,\dots,n$, be the $\nth{i}$ value (for the $\nth{i}$ observation) for the $\nth{j}$ feature.  Let $\mathbf{x}_i=\begin{bmatrix}X_{i,1},\dots,X_{i,m}\end{bmatrix},\:i=1,\dots,n$, be the $\nth{i}$ row vector of feature values for a particular observation $i$; $Y_i$ is the corresponding target feature.

For an independent sample of values $\mathbf{X}$ and some parametric distribution with parameters $\boldsymbol{\theta}$, the likelihood is the product of the density function evaluated at each observation $\mathbf{x}_i\mid\boldsymbol{\theta}$.  For a predictive model, such as linear regression, the likelihood relates the fit of the estimated regression parameters, given input $\mathbf{X}$ (which produces a prediction $\hat{\mathbf{Y}}$) to the true values $\mathbf{y}$.

For linear regression, the parameters are $\boldsymbol{\theta}=(\hat{\boldsymbol{\beta}}, \hat{\sigma}^2)$. $\hat{\boldsymbol{\beta}}$ is an $m$-length vector of regression coefficient estimates of the true values $\boldsymbol{\beta}$, and $\hat{\sigma}^2$ is the mean squared error of $\mathbf{Y}$ versus $\hat{\mathbf{Y}}$. The general equation for regression, relating the inputs $\mathbf{X}=\{X_{i,j}\}$ and $\mathbf{Y}$ is that 

\[
Y_i=\left(\sum_{j=1}^{m}\beta_jX_{i,j}\right) + \epsilon_i
\]
where $\epsilon_i\sim \textrm{iid}\:\:\mathcal{N}(0, \sigma^2)$.  The linear regression estimate (i.e., relation) is 

\[
\hat{Y}_i=\sum_{j=1}^{m}\hat{\beta}_j X_{i,j}
\]

The regression prediction error variance is $\hat{\sigma}^2=\frac{(\mathbf{X}\hat{\boldsymbol{\beta}} - \mathbf{Y})^T(\mathbf{X}\hat{\boldsymbol{\beta}} - \mathbf{Y})}{m}$, where the numerator is the dot product of the prediction residual ($\hat{\mathbf{Y}}=\mathbf{X}\boldsymbol{\beta}$).
Given $\boldsymbol{\theta}$ and inputs $\mathbf{D}=(\mathbf{X},\mathbf{Y})$ (these can be the particular inputs used to fit the model, or other ones, $\mathbf{D}'=(\mathbf{X}', \mathbf{Y}')$, of the same form, such as a field set), the regression likelihood follows from the fact that the prediction errors are normally distributed and independent (a standard assumption of linear regression which we make here):

\[L(\boldsymbol{\theta},\mathbf{X},\mathbf{Y})=\prod_{i=1}^n \frac{1}{\sqrt{2\pi\hat{\sigma}^2}}\exp{\left(-\frac{(\mathbf{x}_i\hat{\boldsymbol{\beta}}-\mathbf{Y}_i)^2}{2\hat{\sigma}^2}\right)}
\]

Taking the log of the above, and factoring out unnecessary constants, we get the log-likelihood:

\[
LL(\boldsymbol{\theta},\mathbf{X},\mathbf{Y})\propto n(\ln(2\pi\hat{\sigma}^2)) + (1/\hat{\sigma}^2)\sum_{i=1}^n(\mathbf{x}_i\hat{\boldsymbol{\beta}}-\mathbf{Y}_i)^2
\]

The dimension of a parametric model is the number of estimated parameters.  Let $\mathcal{L}_h$ now designate a particular linear regression model (i.e., relation).  In the case of a given $\mathcal{L}_h$, this dimension is $d_h=|\boldsymbol{\beta}| + 1$ (number of regression parameters plus 1 for the error variance).  The Bayesian Information Criterion (BIC) for $
\mathcal{L}_h$ on a given input data set $\mathbf{D}=(\mathbf{X},\mathbf{Y})$, is $\textrm{BIC}(\mathcal{L}_h,\mathbf{D})=-2LL(\boldsymbol{\theta}_h,\mathbf{X},\mathbf{Y}) + d_h\ln{(n)}$. The BIC takes into account the fit (via the log-likelihood) in addition to the number of parameters and data set size $n$.  Lower BIC is better; this means a given fit (log-likelihood) achieved with more parameters is worse (i.e., model size is penalized).

Now, given two pairs $(\mathcal{L}_1, \mathbf{D})$ and $(\mathcal{L}_2,\mathbf{D}')$ of model/data input combinations, we can compare the relative fit of each model on its associated data to that of the other pair. In both pairs, the model or data inputs can be the same, or different (i.e., $\mathcal{L}_1=\mathcal{L}_2$ or $\mathbf{D}=\mathbf{D}'$); the first case would correspond to comparing a single model's fit on two different inputs, or two different models' fits on the same inputs, to pick one model over the other.  This comparison can be done by calculating the Bayes factor (BF) for the two pairs, which is approximately 

\[
\textrm{BF} \approx \exp\left(-0.5(\textrm{BIC}(\mathcal{L}_1,\mathbf{D}) - \textrm{BIC}(\mathcal{L}_2,\mathbf{D}'))\right)  
\]

The BF is a standard metric of comparing the fit of two alternative models (here, the polynomial relations) on data sets.  In our application, $\mathcal{L}_1=\mathcal{L}_2$, and so the BF will allow us to decide whether the linear relation $\mathcal{L}_1$'s fit, or strength, has changed significantly between $(\mathbf{X},\mathbf{Y})$ and $(\mathbf{X}',\mathbf{Y}')$.  In particular, $\mathcal{L}$ will be a relation whose parameter estimates $\boldsymbol{\theta}=(\hat{\boldsymbol{\beta}},\hat{\sigma}^2)$ are derived from the first inputs $(\mathbf{X},\mathbf{Y})$.

If the $\textrm{BF} > 1$, this is evidence in favor of $L_1$ fitting $(\mathbf{X},\mathbf{Y})$ better than $(\mathbf{X}',\mathbf{Y}')$.  In contrast, if $0<\textrm{BF} < 1$, the fit on $(\mathbf{X}',\mathbf{Y}')$ is better. 
For a constant $c>1$, if the BF$=c$ or $=1/c$, the larger $c$ is, the more conclusive is the evidence; BFs of $c$ or $1/c$ are the same strength of evidence in favor of the respective decisions. A decision of `drift' of the relation $\mathcal{L}$'s fit between the two input pairs is made if the BF passes some pre-determined threshold $c>1$ or $1/c<1$.  

In particular, since $\mathcal{L}_1$ is determined from $\mathbf{D}$, we expect the BF$>1$, since it should fit the data $\mathbf{D}$ it is derived on better than some other $\mathbf{D}'$.  However, the magnitude of the BF will determine if this drift, or change in fit, is statistically significant.

\subsection{Related work: using multiple Bayes Factors in hypothesis testing}
\label{ssec:multiple_bfs}

In Section~\ref{ssec:BF_for_drift_detection}, we noted we will use Bayes Factors to measure whether the fit of a linear model $\mathcal{L}_j$ changes significantly between two data sets $\mathbf{D}$ and $\mathbf{D}'$.  Specifically, as discussed in Section~\ref{ssec:experimentalDesign}, we will actually use the set of Bayes Factors on a set of relations $\{\mathcal{L}_j\}$, rather than just one relation $\mathcal{L}_j$.
In the field of multiple hypothesis testing, a set of (typically related) statistical hypotheses $\{H_j\}$ are conducted, and it is desired to make a single decision based on the combined results.  For instance, each hypothesis may yield a p-value $p_j$, and we want to make a single decision based on $\{p_j\}$ rather than rejecting each hypothesis individually or not.

While the literature on p-value-based multiple testing is very extensive, it seems that similar use of multiple BFs in a decision is limited.  For instance, \cite{LCLHMGT17} use BFs on genetic multiple markers, rather than a single locus, to measure the association between genetic regions.  This is an extension we will pursue in future work, but for now we simply examine the ability for individual linear relations to detect drift through changes in their BFs.

\section{Experiments}
\label{sec:experiments}
In this section, we describe our experiments. First, we discuss the data sets on which the experiments were conducted, and then we detail the experiment design and results.  

\subsection{Data sets}
\label{ssec:dataAndClassification}
We validate our algorithm on several data sets. Following is a description of the data sets being used in our experiments:
\begin{itemize}
\item Rain in Australia \cite{Rain} is a structured data set which contains daily weather observations from numerous Australian weather stations. It consists 142k records with 15 numerical features.
\item London bike sharing \cite{LondonBike} is a structured data set which contains historical data for bike sharing in London. It consists of 17k records with 12 numerical features.
\item Personal loan modeling \cite{PersonalLoan} is a structured data set which predicts whether a customer will respond to a Personal Loan Campaign. It consists of 5k records with 14 features.
\end{itemize}

\subsection{Experimental Design}
\label{ssec:experimentalDesign}
The  goal  of  the  experiments  is  to  estimate  the effectiveness of the algorithm for identifying drift over the mentioned data sets, simply by measuring changes in the BFs of given linear relations $\{L_j\}$. Using these data sets, we will simulate scenarios where there is a change in the data distribution over time as well as in the influence of existing features over other features. For simplicity, we only generated polynomial relations of maximum degree $\ell=2$ using $k=2$ other data set features, for each target feature $Y$.

In our experiments, the amount of drift is influenced by a controlling parameter of the simulation, whose value we change gradually to incrementally increase the amount of simulated drift, which should be easier to detect. Observations from this new ‘drift data’ should typically cause the linear relations' strengths to change; here that is determined by the BF indicating it is no longer a good fit on the drifted data set. 

First, we randomly split each data set into a baseline data $\mathbf{D}$ and field data set $\mathbf{D}'$ of equal size.  A set of linear relations $\{\mathcal{L}_j\}$ are determined on the baseline data $\mathbf{D}$; each relation's strength is measured by its regression model $R^2$ metric, where a higher value means the estimated relation between $Y$ and the $k$ other selected features is stronger.  

Let $\underline{\mathbf{D}}'$ be the field data $\mathbf{D}'$ after a simulated alteration is performed to induce drift. When no drift is induced, we expect a given relation $\mathcal{L}_j$ to fit roughly equivalently on $\mathbf{D}$ and $\mathbf{D}'$; that is, its BF should be close to 1 when these data sets are compared due to the randomness of the split.  This is analogous to specifying a null hypothesis that two data samples should are distributed the same (e.g., that their difference in means is 0), and a low p-value is used to reject this hypothesis. Similarly, because of the random split, $\mathbf{D}$ and $\mathbf{D}'$ should be approximately distributed the same, so if the BF indicates there is a significant difference in a relation $\mathcal{L}_j$'s fit between them, we can decide there has been drift (i.e., the distribution is not the same) with a certain level of statistical certainty.

A relation $\mathcal{L}_j$ is a good diagnostic sensor of drift if its BF reacts strongly to induction of drift, when $\mathbf{D}$ and the drifted $\underline{\mathbf{D}}'$ are compared, but it does not (BF remains close to 1) when drift is not induced on $\mathbf{D}'$.  Our contention here is that the stronger relations $\mathcal{L}_j$, based on features which are more strongly correlated with the target $Y$ (see Section~\ref{ssec:identifying_relations}), will be better drift sensors than weaker relations.

We conduct two types of drift simulation, described as follows:

\begin{enumerate}     
    \item \textbf{Row permutations}: This technique is described in full in (\citealt{ADFRZ21}, Section III.C), and corresponds to experiment settings $E_3$ with $C=1$.  For each column, a fixed proportion $R\in [0,1]$ (`perc') of rows are randomly selected, and their values permuted within the column.  This is random drift which maintains the typical marginal feature distributions, but disrupts inter-feature associations, which can affect linear relations between them on the baseline data $\mathbf{D}$. If $R=0$, the rows of the full data set are resampled with replacement (should result in very similar marginal and inter-feature distributions), while higher $R>0$ correspond to greater amounts of shuffling of values.   
    \item \textbf{Relation unfairness}: In contrast to the permutation drift, which has non-controlled affects on associations, here we create a distortion $\underline{\mathbf{D}}'$ from $\mathbf{D}'$ by changing linear relations in a targeted way.  This is a synthetic example couched in the context of algorithmic fairness, using the personal loan data set \cite{PersonalLoan}.  There is a growing literature on the principle of algorithmic fairness in ML, in which, say, a certain prediction or decision by an ML model should not be affected by certain `sensitive' features (e.g., a person's race or sex), conditional on other features that are relevant.  For instance, a black family (`RACE' is a a sensitive feature) should not be denied a loan when a white family with the same financial (relevant) characteristics is granted one. However, there may be certain associations between the sensitive and relevant features; for instance, black families may, on average, be less likely to receive loans if black families on average may tend to have lower income, and not because of their race.
    \\\\Here, we use only numeric features and simulate a scenario in which a feature $Y$ (mortgage loan, for which we use the feature for the current mortgage value) should only depend on a subset of relevant features $\mathbf{R}=\{ \textrm{INCOME}, \textrm{CREDIT CARD SPENDING}\}$, but not on certain sensitive features $\mathbf{S}=\{\textrm{AGE}, \textrm{SIZE OF FAMILY}\}$.\\\\We have an unfairness parameter $u \geq 0$.
    When $u=0$ (completely fair), $Y$ should depend only on features in $\mathbf{R}$.
    This is done by generating $\tilde{Y}$, a synthetic version of $Y$, from a linear regression on only the relevant features, with Gaussian noise.  Thus, on two random splits with $u=0$, we have a baseline $\mathbf{D}=(\mathbf{X}, \tilde{\mathbf{Y}})$ and field data $\mathbf{D}'=(\mathbf{X}', \tilde{\mathbf{Y}}')$, where each $\tilde{\mathbf{Y}}$ is synthetically generated on the basis of the original mortgage loan feature. Of course, the synthetic $\tilde{Y}$ may still be \textit{correlated} with features in $\mathbf{S}$, due to their correlations with features in $\mathbf{R}$, but it isn't determined by them, just like the bank loan shouldn't be determined by race, given other features. \\\\
    When $u>0$, the synthetic $\tilde{Y}$ is generated by a regression of the original $Y$ on features in $\mathbf{R}$, and also on the features of $\mathbf{S}$, which are each scaled by $u$.  This is done by pre-standardizing all features, saving the original standard deviations, then rescaling the standardized coefficients for $\mathbf{S}$ by $u$, to generate $\tilde{Y}$ in a similar domain space as the original $Y$; see \cite{G21}.  At each level of $u>0$, only $\tilde{\mathbf{Y}}'$, and not the features $\mathbf{X}'$ in $\mathbf{D}'$, is altered, to generate a drifted version $\underline{\mathbf{D}}'=(\mathbf{X}',\underline{\tilde{\mathbf{Y}}}')$ of $\mathbf{D}'$. Increasing $u$ means the influence of the sensitive features on $\tilde{Y}$ increases; when $u=1$, features in $\mathbf{R}$ and $\mathbf{S}$ should be equally influential on $\tilde{Y}$, while for $u>1$, features in $\mathbf{S}$ have more influence than those in $\mathbf{R}$.  In all cases, $\tilde{Y}$ is centered and scaled to have a fixed mean and standard deviation.\\\\
    Thus, drift detection in this simulation entails first finding strong relations on $\mathbf{D}=(\mathbf{X},\tilde{\mathbf{Y}})$ when $u=0$, and then creating $\underline{\mathbf{D}}'=(\mathbf{X}', \underline{\tilde{\mathbf{Y}}}')$ from $\mathbf{D}'$ for increasing levels of $u$.  Any relation involving the synthetic mortgage target $\tilde{Y}$ and features in $\mathbf{R}$ and $\mathbf{S}$ should tend to change as $u$ changes, because $\tilde{Y}$ is synthetically generated in a different way depending on $u$.  We expect that BFs for these relations will change with $u$.
\end{enumerate}

\begin{figure}
 \includegraphics[width=0.47\linewidth]{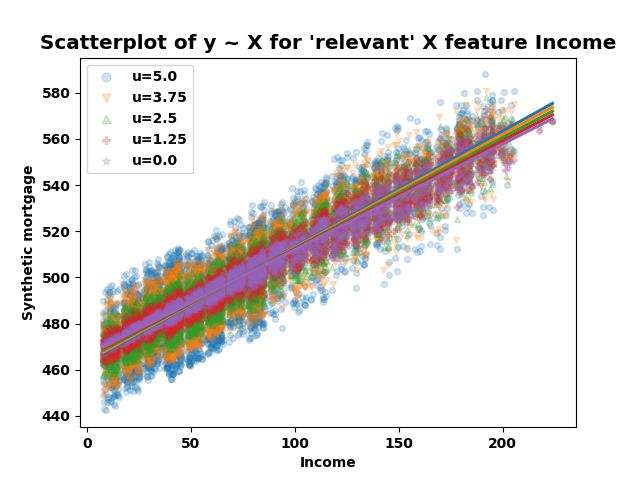}
       \includegraphics[width=0.47\linewidth]{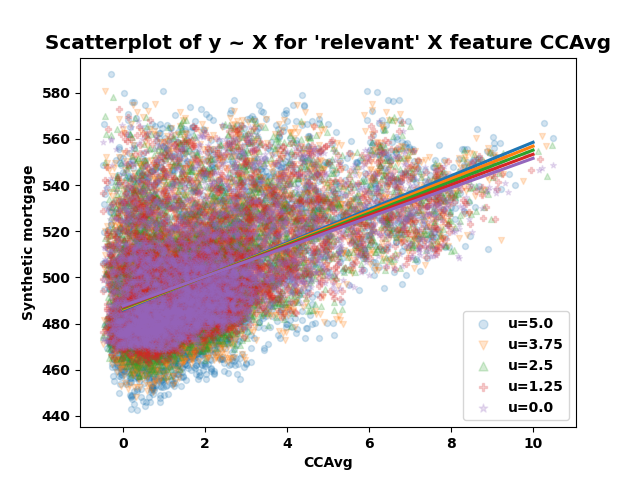}
       \caption{Synthetic mortgage vs relevant features ($\mathbf{R}$), for increasing unfairness $u$.}
     \label{fig:fair_scatter} \includegraphics[width=0.47\linewidth]{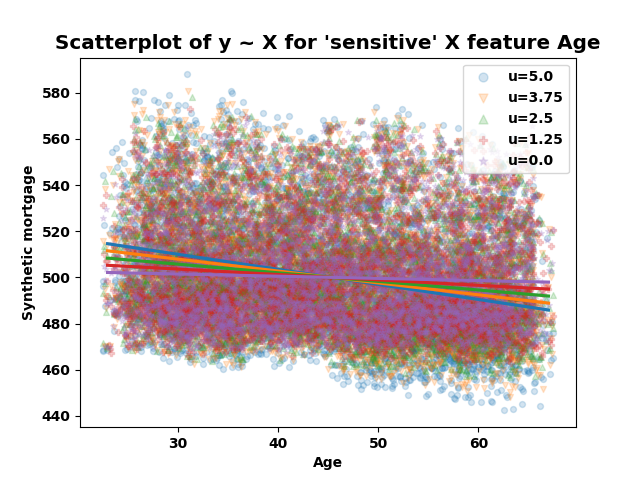}
       \includegraphics[width=0.47\linewidth]{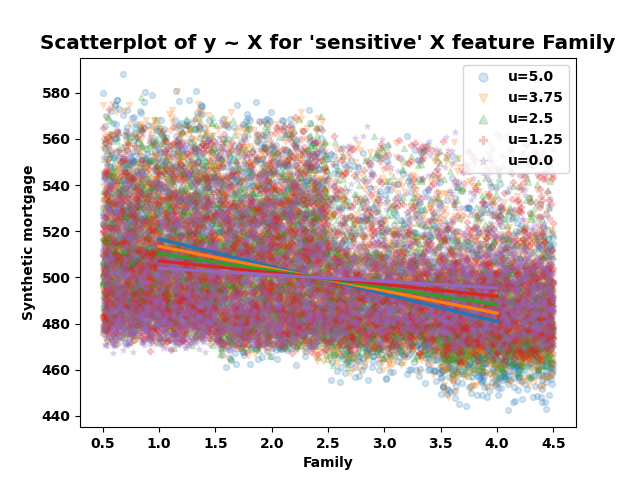}
       \caption{Synthetic mortgage vs sensitive features ($\mathbf{S}$), for increasing unfairness $u$.}       \label{fig:unfair_scatter}
\end{figure}

\begin{figure}
       \includegraphics[width=0.47\linewidth]{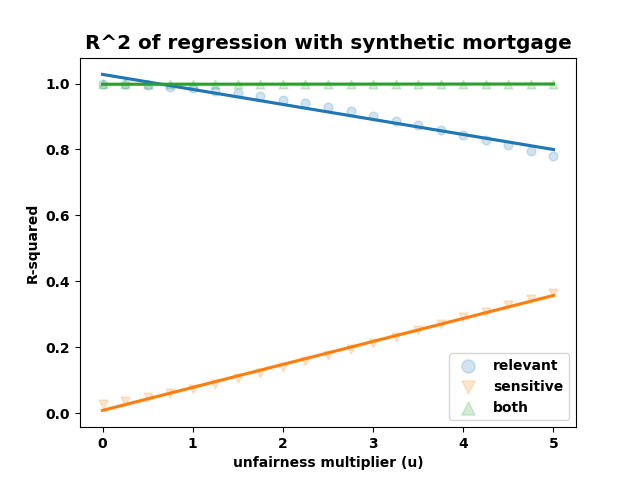}
       \includegraphics[width=0.47\linewidth]{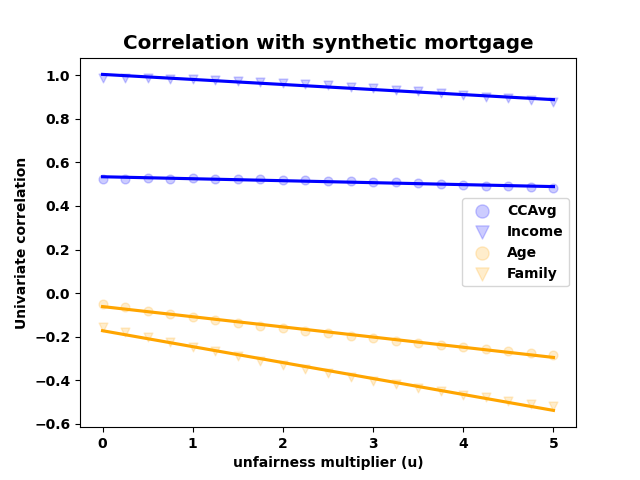}
       \caption{Regression $R^2$ and univariate correlation for increasing unfairness $u$.  In the right panel, the orange features belong to $\mathbf{S}$.}       \label{fig:fairness_correlations}
\end{figure}

We illustrate some examples of drift insertion for type 2, where the relevant and sensitive feature sets are $\mathbf{R}$ and $\mathbf{S}$ as above.  Figures~\ref{fig:fair_scatter} and \ref{fig:unfair_scatter} show scatterplots of the synthetic target vs each of the regression features.  Increasing $u$ should increase the influence of features in $\mathbf{S}$, and decrease the influence of features in $\mathbf{R}$, on $\tilde{Y}$.  Although, for instance, the \textit{slop}e of the regression line of $\tilde{Y}$ on the relevant feature INCOME actually increases with $u$, Figure~\ref{fig:fairness_correlations}, the effect is that for increases in $u$, the \textit{regression fit} $R^2$ decreases for $\mathbf{R}$ and increases for $\mathbf{S}$, and that the correlations of $\tilde{Y}$ decrease in absolute value for features in $\mathbf{R}$, while increasing for features in $\mathbf{S}$.  These correspond with a given relation between $\tilde{Y}$ and $\mathbf{R}$, which may be very strong initially (high $R^2$ when $u=0$) losing predictive power with increasing $u$; the reverse is true for relations involving $\mathbf{S}$.

In either of the two ways of inducing drift discussed above, our experiments will be based on the following procedure: First, $\mathbf{D}=(\mathbf{X}, \mathbf{Y})$ is the baseline data set, on which a given relation $\mathcal{L}_j$ is identified. $\mathbf{D}'=(\mathbf{X}', \mathbf{Y}')$ is a data set that should theoretically follow the same distribution as $\mathbf{D}$, derived from a random data set split.  The BF is to compare the fit of $\mathcal{L}_j$ on $\mathbf{D}$ and $\mathbf{D}'$ is calculated.  In this case, a false positive occurs if, on the basis of the BF, we decide that statistically significant drift has occurred, when in fact, it shouldn't.  We will typically expect the BF to be close to 1, and not statistically significant, which demonstrates the BF's ability to control the false positive rate, that is, having a low known probability of making a false determination of drift.

Then, we will induce drift in $\mathbf{D}'$ by using the simulations described in Section~\ref{ssec:experimentalDesign}, thus creating a new data set $\underline{\mathbf{D}}'$.  The BF will then be calculated to compare $\mathcal{L}_j$ on the original $\mathbf{D}$ versus the drifted $\underline{\mathbf{D}}'$. When the amount of drift (which we control via a simulation parameters) increases, we expect the BF to increase until at some point we are able to detect statistically significant drift in the data sets, simply by examining changes in the relation's BF.

As mentioned before, for strong relations $\mathcal{L}_j$ (with high $R^2$ values) we expect to see a growth in the Bayes Factor value when drift is introduced (through the shuffling of values or distortion by increased unfairness $u$). In contrast, for weak relations the calculated Bayes Factor is not supposed to be affected in a monotonic way by the drift, and thus is less useful in detecting the drift.  

These drift induction approaches significantly change the data distribution and the relations between features in the field data $\underline{\mathbf{D}}'$.  Ultimately, the goal of this work is to show that we can use linear relations to detect drift in data sets.  If we have a trained any predictive ML model (not the relations) on a baseline data set, it is likely its performance will degrade on a drifted dataset (e.g., $\underline{\mathbf{D}}'$), which we hope to detect in advance by changes in the fit of the chosen linear relations $\{\mathcal{L}_j\}$.

\subsection{Experimental Results}
\label{ssec:experimentalResults}

In Section~\ref{ssec:experimentalDesign} we suggested a way to estimate the effectiveness of the approach for identifying drift and used it over the data sets mentioned in Section~\ref{ssec:dataAndClassification}. We calculated the potential relations, and partitioned them to strong and weak relations as mentioned earlier. The next step was a random split of the data sets, and introduce of a drift to the field data by shuffling the rows. 

\begin{figure}
       \includegraphics[width=0.47\linewidth]{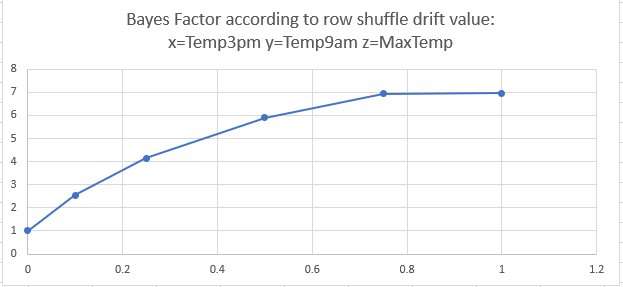}
       \includegraphics[width=0.47\linewidth]{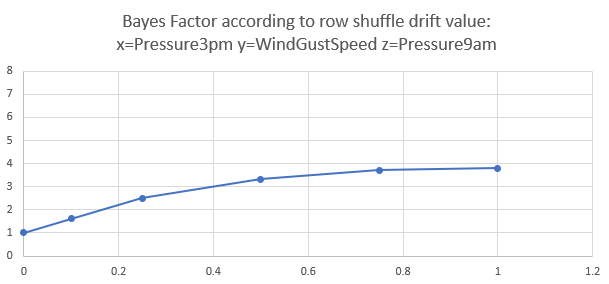}
       \caption{Bayes Factor according to row shuffle drift value for strong relations}
     \label{figure:BF_shuffle_strong}
\end{figure}

\begin{figure}
       \includegraphics[width=0.47\linewidth]{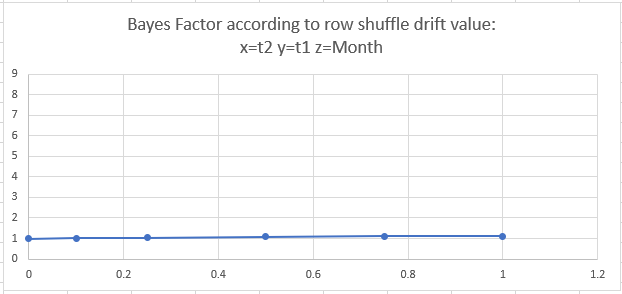}
       \includegraphics[width=0.47\linewidth]{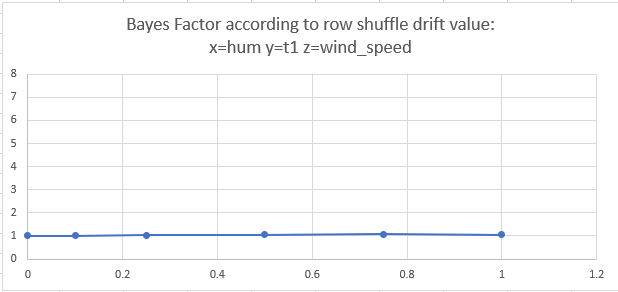}
       \caption{Bayes Factor according to row shuffle drift value for weak relations}
     \label{figure:BF_shuffle_weak}
\end{figure}

Figure~\ref{figure:BF_shuffle_strong} shows the results for a selected strong relation. Specifically, the selected relations has $R^2$ metric value which is very high. For these relations, the Bayes Factor value was rapidly changed due to the row shuffle drift value. In contrast, Figure~\ref{figure:BF_shuffle_weak} shows the results for weak relations with low $R^2$ metric. For these relations, the Bayes Factor value was always around 1 (no effect on the relation's shift when the drift increases). 

This presents the ability to identify a drift in the data according to the Bayes Factor value calculated over the strong relations. Moreover, it demonstrates a clear difference between the influence of strong relations for the drift identification in compare to the weak relations.

\begin{figure}
       \includegraphics[width=0.47\linewidth]{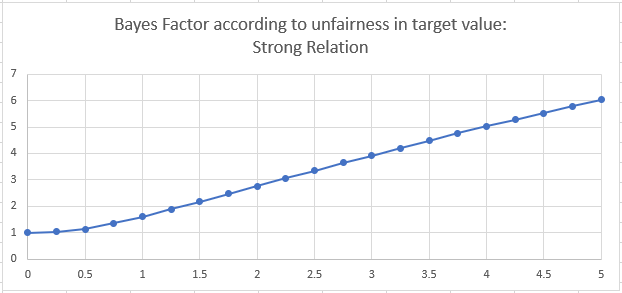}
       \includegraphics[width=0.47\linewidth]{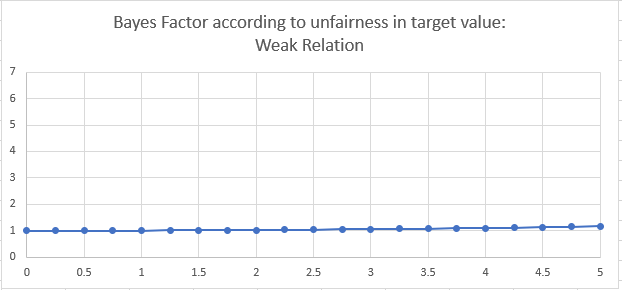}
       \caption{Fairness experiment results of Bayes Factor increased due to unfairness}
    \label{figure:BF_unfair}
\end{figure}

For the relations unfairness experiment, we took the Personal Loan Modeling data set from Section~\ref{ssec:dataAndClassification} and calculated a relation for the synthetic `MORTGAGE' target feature. As mentioned in Section~\ref{ssec:experimentalDesign}, we partitioned the data set to baseline and field, introduced increased unfairness to the field data, and calculated the Bayes Factor. The features selected as correlated for the relation were `INCOME' and `CCAvg'. The experiment results shows in Figure~\ref{figure:BF_unfair} that the Bayes Factor value was rapidly increased according to the amount of unfairness introduced. We also tried to predict the target feature by the features `AGE' and `EXPERIENCE' which made a weak relation (low $R^2$ metric value). In this case we show that the Bayes Factor was almost not changed.

\section{Discussion
}
\label{sec:discussion}

We report on initial promising results of detecting drift in the data distribution using polynomial relations.

In our experiments we demonstrated a strong connection between the drift and the ability to detect it via the proposed method of polynomial relations. We plan to extend this to additional drift types and more types of potential relations. We believe that our technology has other useful use cases. One example for a use case may be comparing two existing data sets (with a similar structure) and returning a result which indicates the similarity of these data sets.

There are additional challenges in improving the polynomial relations approach. One important challenge is an improvement of the heuristic that selects the features which participate in the relation. Currently, our heuristic works well mainly with features that are directly correlated. We plan to extend it for cases in which several features are only correlated together to a target feature. For example, the relation $X_1-X_2=X_3$, where $X_1$ and $X_2$ are not individually highly correlated to $X_3$, but the function $X_1-X_2$ is highly correlated with $X_3$.

Another important challenge is the trade-off between the complexity of the relation, over-fitting, understanding the relation, and the ability to detect drift. We plan to experiment with techniques to address that. One example is the selection of the right degree $\ell$ of the polynomial relation. Improving the heuristic and accounting for the various trade-offs should result in a faster and more accurate detection of drift.

Also, we plan to experiment with using LASSO with a regularization penalty, rather than OLS linear regression, to determine the optimal relation on a given embedding $\mathcal{P}(\mathbf{v})$ when $\mathbf{v}$ is the set of most correlated other features with the target $Y$.  If the maximal number of features $k$ or maximal degree $\ell$ are larger, the embedding $\mathcal{P}$ will grow exponentially, likely causing over-fitting.  LASSO can decide which embedding elements to actually include in the relation.  Furthermore, we plan to see what is the effect of standardizing the embedding features before determining the relation.

Lastly, following the method in \cite{LCLHMGT17}, we want to see whether we can use multiple hypothesis methods to decide on the basis of multiple relations' Bayes Factors, if the entire data set seems to have drifted, instead of examining whether each relation individually has changed when deciding if the data set has drifted.

\section{Conclusion
}
\label{sec:conclusion}
In this work, we addressed the challenge of detecting data distribution drift which may influence the performance of ML models. We automatically identified relations between data set features and statistically detected significant changes in the data which may degrade the model performance if applied on the drifted field data set. We simulated a change in the data over time, and demonstrated the ability to identify it according to the Bayes Factor value calculated over the strong relations.

\bibliography{main}

\end{document}